\documentclass[letterpaper, 10 pt, conference]{ieeeconf}  
\IEEEoverridecommandlockouts     
\usepackage{graphicx}
\usepackage{cite}
\usepackage{picinpar}
\usepackage{amsmath}
\usepackage{url}
\usepackage[latin1]{inputenc}
\usepackage{colortbl}
\usepackage{soul}
\usepackage{multirow}
\usepackage{pifont}
\usepackage{color}
\usepackage{alltt}
\usepackage{hyperref}

\usepackage{enumerate}
\usepackage{siunitx}
\usepackage{breakurl}
\usepackage{epstopdf}
\usepackage{pbox}
\usepackage[ruled,linesnumbered]{algorithm2e}
\usepackage{booktabs}

\usepackage{bm}
\usepackage{hyperref}

\title{\LARGE \bf
Path Planning with Automatic Seam Extraction over\\Point Cloud Models for Robotic Arc Welding
}

\author{Peng Zhou, Rui Peng, Maggie Xu, Victor Wu and David Navarro-Alarcon
\thanks{All authors are with The Hong Kong Polytechnic University, Kowloon, Hong Kong. Corresponding author e-mail: {\texttt{\small dna@ieee.org}}}%
\thanks{This work is supported in part by the Research Grants Council of HK (grant 14203917) and in part by the Chinese National Engineering Research Centre for Steel Construction Hong Kong Branch (grant BBV8).}%
}

\begin{document}
\maketitle
\thispagestyle{empty}
\pagestyle{empty}

\begin{abstract}
This paper presents a point cloud based robotic system for arc welding. Using hand gesture controls, the system scans partial point cloud views of workpiece and reconstructs them into a complete 3D model by a linear iterative closest point algorithm.
Then, a bilateral filter is extended to denoise the workpiece model and preserve important geometrical information.
To extract the welding seam from the model, a novel intensity-based algorithm is proposed that detects edge points and generates a smooth 6-DOF welding path. 
The methods are tested on multiple workpieces with different joint types and poses. Experimental results prove the robustness and efficiency of this robotic system on automatic path planning for welding applications.
\end{abstract}

\section{Introduction}
Industrial welding robots have been extensively studied with the aim to solve labor shortage problem and improve productivity. Currently, ``teach and playback" mode or offline programming prevails in robotic welding workshops. As these conventional practices lack adaptation to accommodate changes in a working environment, many advanced techniques such as seam detection and tracking are gradually being employed in industry.

Solutions to intelligent robotic arc welding primarily require the assistance of computer vision. 
Previous studies have conducted in-depth research on image processing and established viable methods across several visual sensors, including mono camera, stereo camera, and CCD camera \cite{xu2012real, fan2019initial, diao2017passive, li2017automatic, shah2016review, rout2019advances}. 
\cite{nele2013image} developed an automatic image acquisition system for real-time weld pool analysis and seam tracking. 
The seam characteristics were measured by defining region of interest and implementing pattern learning algorithm. 
\cite{ding2016line} proposed a shape-matching algorithm for an online laser-based seam detection. 
The algorithm enabled autonomous detection for different groove types and localizes the boundary of weld seam. 
\cite{manorathna2014feature} measured the cross-sectional areas of 2D profiles and finds joint features and volume of the workpiece.

While above mentioned studies can effectively locate the seam, their implementations are only validated on simple or flat welding workpieces with common joint types. In most cases, the groove problem is regarded as two-dimensional and image processing is sufficient to identify the target. However, it is important to discuss the generalization of these techniques to cope with complex geometries. The remaining challenge lies in capturing free-form grooves. Point cloud processing is favored as it can comprehensively describe complex 3D shapes. A captured point cloud model can reveal useful geometric information but simultaneously introduce considerable noise. The complexity and computational cost also evolve as the information increases. Therefore, designing a robust algorithm that can efficiently utilise 3D information will be crucial for this application. Although, some recent studies using point cloud processing were able to detect the groove, they did not further exploit the features to improve system functionality, such as plan a welding path with well-defined orientation \cite{zhang2018point, ahmed2018edge, patil2019extraction, jing2016rgb, zhang20193d}. More experiments on complex and diverse seam patterns are needed for verification. In addition, the authors found that most systems need to manually set or program the initial poses for visual sensors to capture the workpiece. This guidance of scanning may be realized in a more intuitive way to facilitate human machine collaboration.

To address the above problems, we propose a novel point cloud based system for robotic welding planning. This system integrates a single RGB-D camera with a cooperative industrial manipulator to fulfill the generic requirements for seam detection. 
Guided by hand gestures, the system captures a set of partial point clouds of the workpiece from different perspectives. 
To reconstruct the surface, it adopts a linear iterative closest point algorithm that improves the rate of fitness and extends bilateral filter algorithm for 3D surface denoising. A novel algorithm is implemented to identify the groove based on edge density. After the localization, the system plans a smooth consecutive welding path with 6-DOF and executes it with the robot manipulator. To validate the proposed methods, we conduct detailed experiments on workpieces with different joint types.

The remaining parts of this article are as follows. Section \uppercase\expandafter{\romannumeral2} introduces the architecture of the proposed planning system and the key techniques applied. The experimental results of the proposed system are presented in Section \uppercase\expandafter{\romannumeral3}. Section \uppercase\expandafter{\romannumeral4} gives final conclusions.

%
%To do:
% Introduction, Logic checking, contribution.
% Compress, add voxel sampling, (compressing mainly focus on )
% Add comparison with STOA (remember to change the workpieces)
% Video:
%

\begin{figure*}[ht]
\vspace{0.2cm}
	\centering
	\includegraphics[width=\columnwidth * 2
	]{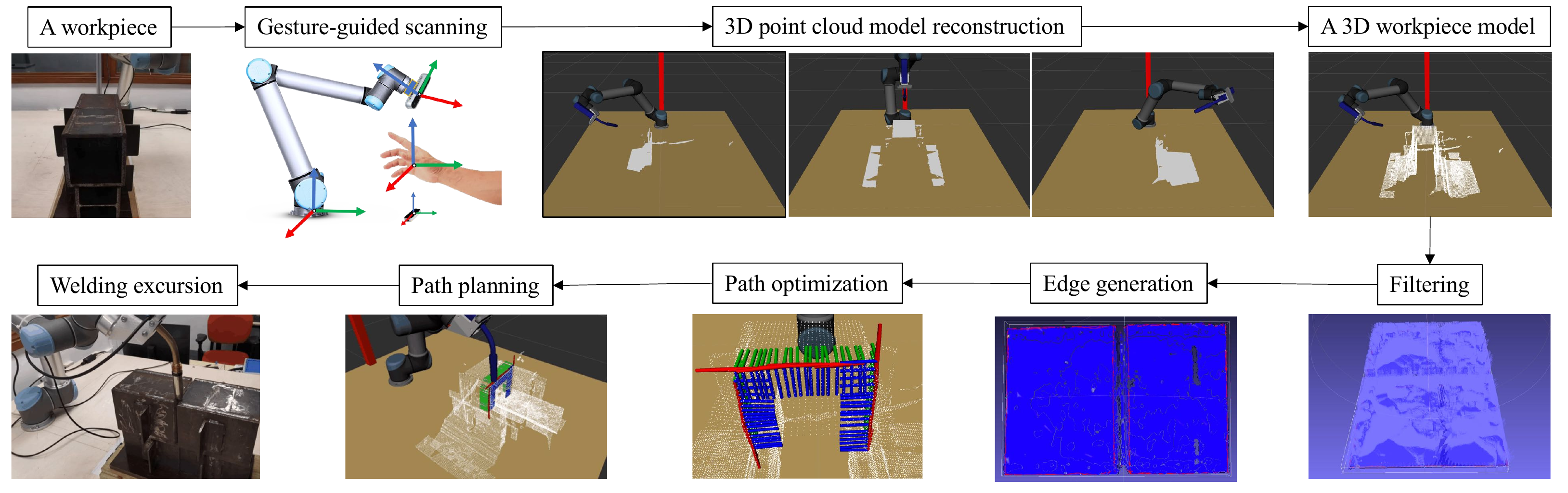}
	\caption{Overview of the automatic robotic welding pipeline. Given a workpiece, a gesture-guided scanning is executed to merge captured point clouds into a completed 3D model, which is further improved with the bilateral filter. Subsequently, on basis of the edge points detected using an edge intensity gradient-based algorithm, a 6-DOF welding path is established.}
	\label{fig:overview}
%	\vspace{-0.2cm}
\end{figure*}

%	Gesture scanning, ICP reconstruction, down sampling, denoising, edge, path planning
%

\section{Methodology}
\subsection{System Overview}
The robotic system architecture is depicted in Fig. \ref{fig:overview}. The complete surface model reconstruction for the target welding workpiece, which a hand gesture-based method, is designed to interact with the manipulator with assistance from a human operator.
Subsequently, the depth camera, which is mounted on the end-effector of the robot, moves to several poses to capture different views of the point cloud.
Then, the point cloud model of the workpiece is reconstructed by combining all the individual frames of the point cloud for the workpiece.
On the basis of the complete surface model, the gradient of the edge intensity-based welding seam detection algorithm is employed to locate the targeted weld path points.
With these generated path points, an interpolation of both translation and rotation can be used to generate a 6-DOF welding path, which can then be sent to the robot manipulator.
Finally, the robot executes the path tracking motion to finish the welding tasks on the targeted workpiece.

\begin{figure}[h]
	\centering
	\includegraphics[width=\columnwidth]{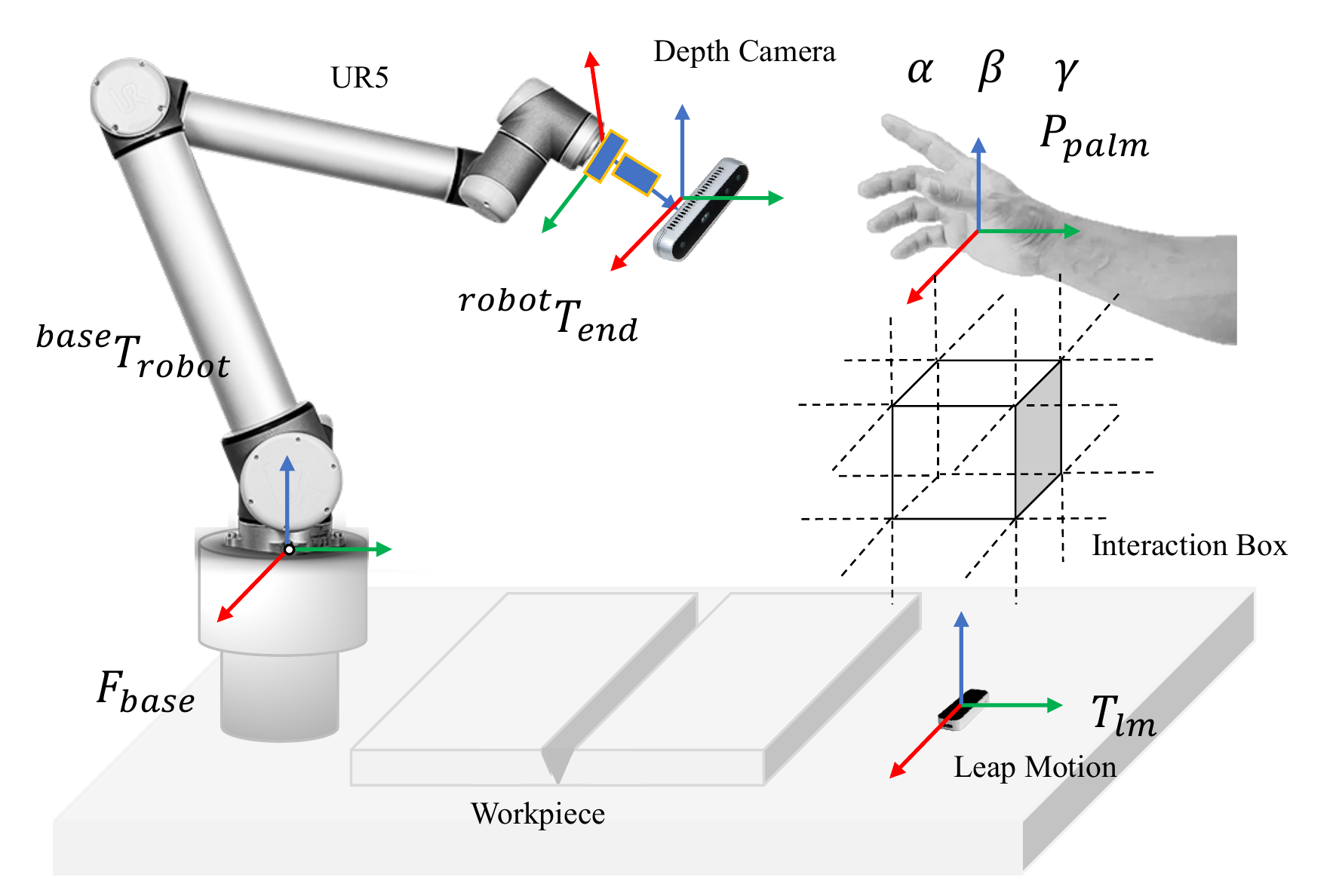}
	\caption{Conceptual representation of the intuitively interactive scanning with hand gestures, in which an interaction box is built to define different mapping rules for robotic motions.}
	\label{fig:gestureControl}
%	\vspace{-0.2cm}
\end{figure}

\subsection{Hand Gesture Guided Scanning}
To scan the welding workpiece, hand gestures are captured for generating control commands to move the camera. As shown in Fig. \ref{fig:gestureControl}, we used a real-world setup to best represent the commonly used cases in this research. A ``joystick" interaction mode is designed to interact with a robot, on which a depth camera is mounted. 
This interaction process is handled by the virtual interaction box above the hand gesture sensor. The moving direction of the scanning depth camera is determined according to the left hand's relative position with respect to the interaction box, and the rotation of the camera is mapped with the raw, pitch and yaw of left-hand gesture captured with leap motion. Meanwhile, the gesture of the right hand is used to control the point cloud capture and the pose of the robot end-effector at the current poses for the succeeding reconstruction process.

\begin{algorithm}[htbp]
	\caption{Hand Gesture Interaction}
	\label{algo:handControl}
	\LinesNumbered
%	\KwIn{Hand Frame $\mathcal{F}_{t}^{hand}$ at time step $t$}
%	Palm position $P_{palm}$, palm Euler angles $(\alpha, \beta, \gamma)$, \\ current pose $ ^{base}\mathbf{p}_{camera}^{t}$ \\
%	\KwOut{Camera pose $\mathbf{p}_{t+1}^{camera}$ at time step $t+1$ }
	\If{True}
	{$(P_{t}^{left}, \alpha, \beta, \gamma)= getPalmAttribute(\mathcal{F}_{t}^{hand})$\\
%		$\mathcal{F}_{t} = LMController.frame(t)$ \\
		$\mathbf{p}_{t}= (P_{t}^{c}, r_x, r_y, r_z) = getPose(t)$ \\

		$r_t = getRegion(P_{t}^{left})$\\
		\If{$r_t$ not in home region}
		{
		$ P_{t+1}^{camera}$ = $\bm{u}_d \cdot l $  \\
		$\mathbf{p}_{t+1}^{camera} = (P_{t+1}^{camera}, r_x, r_y, r_z)$ \\
		\Return pose $\mathbf{p}_{t+1}^{camera}$	
		}
		\Else
		{
			$ ^{lm}\mathbf{R}_{hand} = \text{Transform}(\alpha, \beta, \gamma) $ \\
			$ ^{base}\mathbf{R}_{hand} = {^{base}\mathbf{R}_{lm}} \cdot {^{lm}\mathbf{R}_{hand}}  $ \\
			$ (r_x^{t+1}, r_y^{t+1}, r_z^{t+1})=$ InvRodrigues$(^{base}\mathbf{R}_{hand}) $ \\
			$\mathbf{p}_{t+1}^{camera} = (P_{t}^{camera}, r_x^{t+1}, r_y^{t+1}, r_z^{t+1})$ \\
			\Return $\mathbf{p}_{t+1}^{camera}$
		}
	}
\end{algorithm}

We present the entire interaction process in Alg. \ref{algo:handControl}. In the frame of the UR manipulator, a standard pose is represented with $\mathbf{p} = (x, y, z, r_x, r_y, r_z)$. The algorithm takes the position of the left hand's palm position $P_{palm}$, the hand's Euler angles $(\alpha, \beta, \gamma)$, and the current pose with respect to the camera $ ^{base}\mathbf{p}_{camera}^{\circ}$ as the inputs. The resulting desired output is target pose $^{base}\mathbf{p}_{camera}^{*}$ in the next time step. The entire interaction is in a loop of an infinite frame set $\left\{\mathcal{F}_i, i = t \right\}$ with respect to time step $t$ captured by leap motion sensor. In each frame $\mathcal{F}_i$, the left hand's palm position $P_{palm}$ and its hand's Euler angles $(\alpha, \beta, \gamma)$ are mapped into motions of the robotic manipulator based on the region in which the hand stays with respect to the interaction box above the sensor. According to the relative position with respect to the interaction box, the entire interaction space is separated into seven regions, namely, home, left, right, up, down, inward, and outward regions. When $P_{palm}$ reaches other regions except the home region, 
a corresponding unit vector $\bm{u}_d$ is computed to update the position part of the target pose $^{base}\mathbf{p}_{camera}^{*}$. However, if $P_{palm}$ stays in the home region, the manipulator switches into rotation mode, then the Eule angles captured by the gesture sensor are to be transformed into rotation matrix $^{LM}\mathbf{R}_{hand}$, then this rotation matrix will be left multiplied by a rotation matrix from the gesture sensor to the robot base coordinate system ${^{base}\mathbf{R}_{LM}}$. With the resulting rotation matrix ${^{base}\mathbf{R}_{hand}}$, the inverse $Rodrigues$ formula is used to transform it into the corresponding rotation part $(r_x, r_y, r_z)$ for the target pose. Finally, the orientation part of the target pose $^{base}\mathbf{p}_{camera}^{*}$ is updated. During the process, the right hand's gesture is used to store the point cloud data $\mathcal{P}_i$ and the corresponding pose of robot end-effector $^{base}\mathbf{p}_i$ corresponding to the base frame when a certain ``ok" gesture is detected.

\subsection{Welding Workpiece Reconstruction}
In performing a high precision reconstruction for the welding workpieces, an initial reconstruction is executed on the basis of the scanned point cloud set, $\{\mathcal{P}_1, \mathcal{P}_2, ..., \mathcal{P}_n\}$, and their calculated camera pose, $\{\mathbf{p}_1, \mathbf{p}_2, ..., \mathbf{p}_n\}$, and then a linear iterative Closest point (ICP) algorithm is employed to further improve the fitness performance of the workpiece model. 
For each point cloud $\mathcal{P}$, the transformation from camera coordinate frame with respect to base the frame, $^{b}\mathbf{F}_p$, is given by:
\begin{equation}
{^{b}\mathbf{F}_p} = {^{b}\mathbf{M}_{e}} \cdot {^{e}\mathbf{M}_{c}} \cdot {^{c}\mathbf{F}_p}	
\label{equ:1}
\end{equation}
where ${^{e}\mathbf{M}_{c}}$ is a fixed rigid transformation obtained by calibrating the relative position of the robot end-effector with respect to the origin of the RGB-D camera by means of reverse engineering, and ${^{b}\mathbf{M}_{e}}$ can be easily computed using the inverse kinematics module. To simplify, the origin of the global coordinate system is set at the base of the robot. 

\begin{figure}[htbp]
	\centering
	\includegraphics[width=\columnwidth]{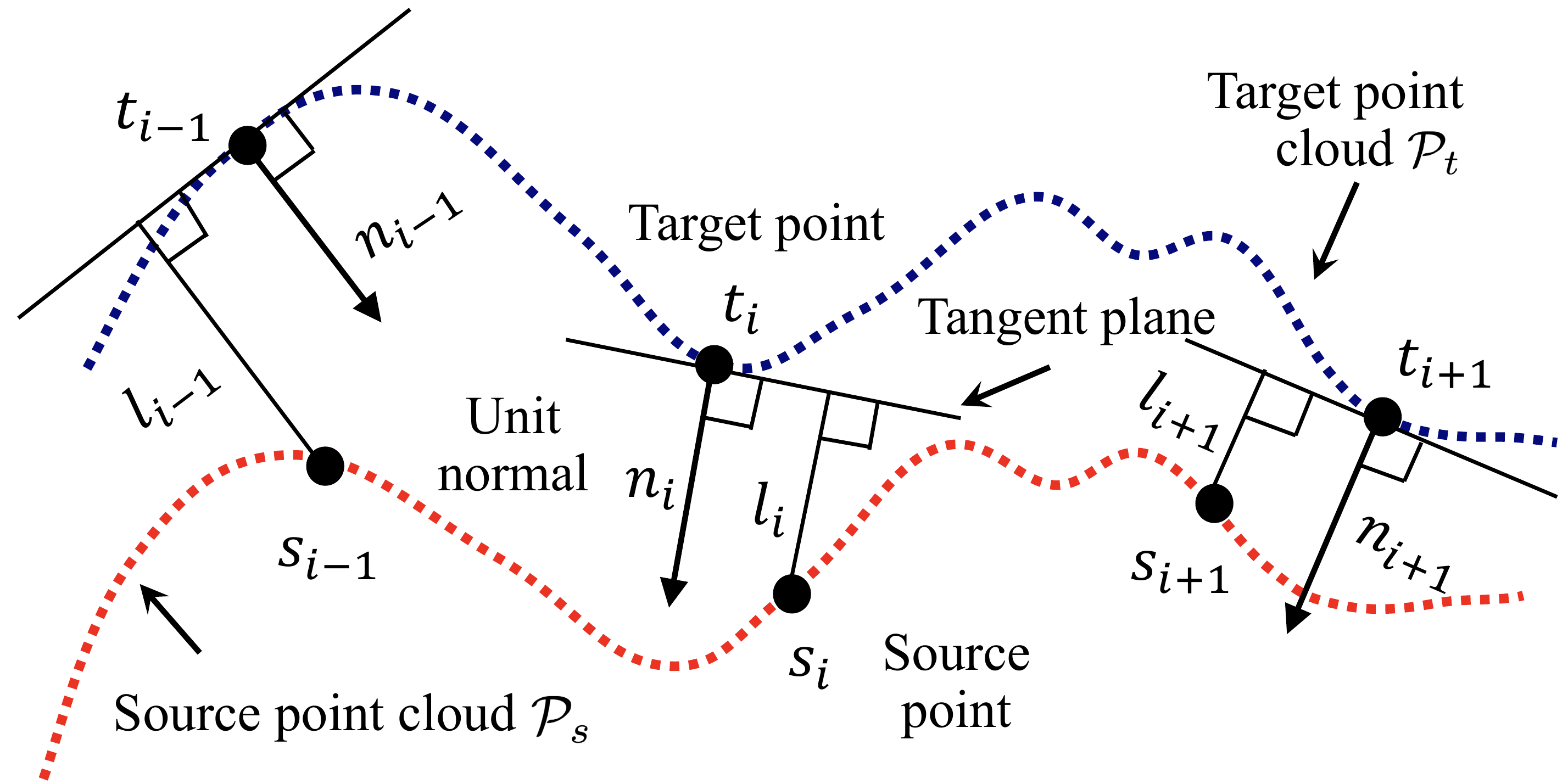}
	\caption{Conceptual representation of the ICP process from a source point cloud $\mathcal{P}_i$ to a target point cloud $\mathcal{P}_t$.}
	\label{fig:ICP}
\end{figure}

By performing the transformations illustrated in Eq. (\ref{equ:1}) on the dataset $\{\mathcal{P}_1, \mathcal{P}_2, ..., \mathcal{P}_n\}$, the entire point cloud data has been set in the base coordinate system, then they can be combined as a completed point cloud $\{{^{b}\mathcal{P}_1} \cup {^{b}\mathcal{P}_2} \cup ... \cup {^{b}\mathcal{P}_n}\}$ for the welding workpiece. In further improving the fitness, an ICP process setting ${^{b}\mathbf{M}_{c}}$ as its initial configuration is needed. 
Specifically, the $i$-th source point is defined as $\mathbf{s}_{i}=\left(s_{i x}, s_{i y}, s_{i z}, 1\right)^{\mathrm{T}}$, its corresponding target point is represented by $\mathbf{t}_{i}=\left(t_{i x}, t_{i y}, t_{i z}, 1\right)^{\mathrm{T}}$, and the unit normal vector at $\mathbf{t}_i$ can be denoted as $\mathbf{n}_{i}=\left(n_{i x}, n_{i y}, n_{i z}, 0\right)^{\mathrm{T}}$. The goal of each ICP iteration is to obtain $\mathbf{M}_{*}$, that is:
\begin{equation}
	\mathbf{M}_{*}=\arg \min _{\mathbf{M}} \sum_{i}\left(\left(\mathbf{M}_{\circ} \cdot \mathbf{s}_{i}-\mathbf{t}_{i}\right) \bullet \mathbf{n}_{i}\right)^{2}
\label{equ:2}
\end{equation}
where $\mathbf{M}_{*}$ and $\mathbf{M}_{\circ}$ are 4$\times$4 3D target and initial transformation matrices. In its first iteration, ${^{b}\mathbf{M}_{c}}$ is set as an initial transform matrix to accelerate the convergence. Usually, a rotation matrix $\mathbf{R}(\alpha, \beta, \gamma) $ and a translation matrix $\mathbf{T}(p_x, p_y, p_z)$ can be decomposed into a 3D transformation $\mathbf{M}$, which can be denoted as:
\begin{equation}
	\mathbf{M}=\mathbf{T}\left(p_{x}, p_{y}, p_{z}\right) \cdot \mathbf{R}(\alpha, \beta, \gamma)
\end{equation}

Eq. \ref{equ:2} is a essentially a least-square optimization problem whose solution normally requires the determination of the six parameters of $\alpha, \beta, \gamma, p_x, p_y$, and $p_z$. However, in the rotation matrix $\mathbf{R}$, three of them ($\alpha, \beta$ and $\gamma$) correspond to the arguments of nonlinear trigonometric functions. Consequently, we cannot employ an efficient linear least-square technique to solve this problem. Nonetheless, the initial iteration ${^{b}\mathbf{M}_{c}}$ can ensure that three angles are $\alpha, \beta, \gamma \approx 0$. Therefore, the approximations $ \sin \theta \approx \theta$ and $\cos \theta \approx 1 $ can change the original nonlinear least-square problem into a linear one. In this manner, a linear least-square algorithm \cite{low2004linear} can still be carried out as follows:

\begin{equation}
%	\mathbf{R}(\alpha, \beta, \gamma)
	\mathbf{\hat R}(\alpha, \beta, \gamma)
 \approx\left(\begin{array}{cccc}1 & -\gamma & \beta & 0 \\ \gamma & 1 & -\alpha & 0 \\ -\beta & \alpha & 1 & 0 \\ 0 & 0 & 0 & 1\end{array}\right)
% =\mathbf{\hat R}(\alpha, \beta, \gamma)
\end{equation}

\begin{equation}
	{\mathbf{\hat M}}=\mathbf{T}\left(p_{x}, p_{y}, p_{z}\right) \cdot \mathbf{\hat R}(\alpha, \beta, \gamma)
\end{equation}

\begin{equation}
	{\mathbf{\hat M}_{*}}=\arg \min _{\hat{\mathbf{M}}} \sum_{i}\left(\left(\hat{\mathbf{M}} \cdot \mathbf{s}_{i}-\mathbf{t}_{i}\right) \bullet \mathbf{n}_{i}\right)^{2}
\end{equation}
where $\left(\hat{\mathbf{M}} \cdot \mathbf{s}_{i}-\mathbf{t}_{i} \right)$ can be written as a linear expression with six parameters ($\alpha, \beta, \gamma, p_x, p_y$, and $p_z$). Then, $N$ pairs of point correspondences can be arranged as a linear matrix expression as follows:
\begin{equation}
	\min _{\hat{\mathbf{M}}} \sum_{i}\left(\left(\hat{\mathbf{M}} \cdot \mathbf{s}_{i}-\mathbf{t}_{i}\right) \bullet \mathbf{n}_{i}\right)^{2}=\min _{\mathbf{x}}|\mathbf{A} \mathbf{x}-\mathbf{b}|^{2}
\end{equation}
Thus, $\hat{\mathbf{M}}_{*}$ can be obtained by solving: 
\begin{equation}
	\mathbf{x}_{*}=\arg \min _{\mathbf{x}}|\mathbf{A} \mathbf{x}-\mathbf{b}|^{2}
\end{equation}
which can be further solved using singular value decomposition (SVD).

\subsection{Point Cloud Denoising}
In removing noise from the point cloud data without removing vital edge information in our succeeding procedure, the bilateral filter specific for point cloud data is considered. Normally, the bilateral filter is well defined for 2D pixels. Thus, we design and extend a similar version for point cloud data.
Specifically, for a given point $p \in \mathcal{P}$, a unit normal vector $\mathbf{n}_{p}$ is estimated for each point $p$ via its neighbors $\mathcal{N}_{r}(p)= \left\{q \in \mathcal{P} \mid \|q-p\|_{2}<r\right\}$. A regression plane is fitted by the neighbor point sets, then the normal of the plane is regarded the normal of the query point. The normal computation function is established as:
\begin{equation}\label{normal}
	\begin{aligned}
	  \mathbf{n}_p = \arg \underset{\mathbf{n}}{\min} \sum^n_{i=1}((q_i-p_c)^T \mathbf{n})^2
	\end{aligned}
\end{equation}
where $\mathbf{n}_p$ is the targeted normal, $p_c$ is the gravity center point of the current point's neighbor, and $q_i$ is the $i$-th neighbor point within the neighbor. 
By computing the mean and covariance matrix of the neighbors, this least-square regression plane can be estimated. 

\begin{figure}[h]
	\centering
	\includegraphics[width=\columnwidth]{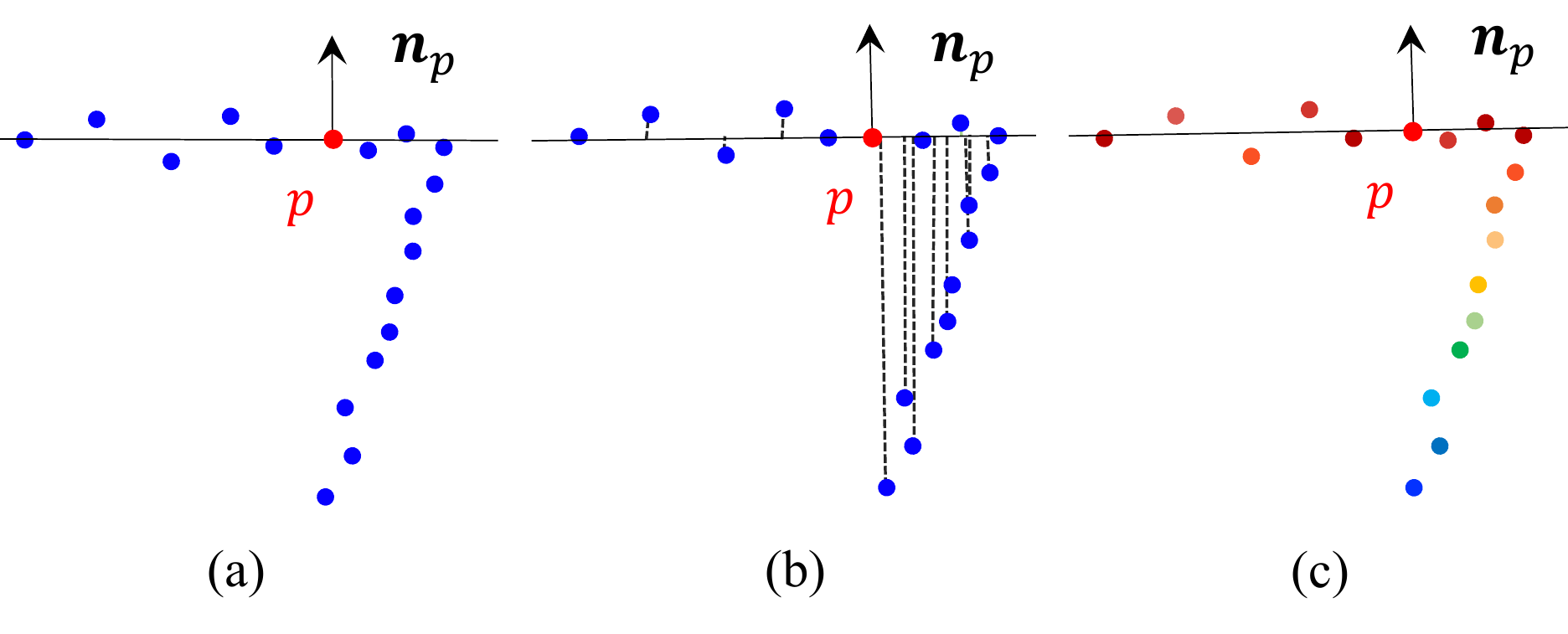}
	\vspace{-0.8cm}
	\caption{Contributions of neighbors to the denoised point $p$ in the bilateral filtering. 
	(a) Regression plane estimation.
	(b) Distance weights.
	(c) Normal distance weights. (Blue: small weights; red: large weights)	
	}
	\label{fig:filter}
%	\vspace{-0.2cm}
\end{figure}

With the estimated normal $\mathbf{n}_p$, $p^{\prime}=p+\Delta p \cdot \mathbf{n}_{p}$ is used to update the point position to $p^{\prime}$, and $\Delta p$ is a normal distance weight defined as:
\begin{equation}
	\Delta p=\frac{\sum_{q \in \mathcal{N}_{r}(p)} w_{m}(\|q-p\|) w_{n}\left(\left|\left\langle\mathbf{n}_{p}, q-p\right\rangle\right|\right)\left\langle\mathbf{n}_{p}, q-p\right\rangle}{\sum_{q \in \mathcal{N}_{r}(p)} w_{m}(\|q-p\|) w_{n}\left(\left|\left\langle\mathbf{n}_{p}, q-p\right\rangle\right|\right)}
\end{equation}
where $w_m$ and $w_n$ are the weights generated by Gaussian distributions with variances $\sigma_m$ and $\sigma_n$, respectively. Fig. \ref{fig:filter} shows the conceptual representation of the process of filtering. In this scheme, the weights on the distances will be balanced by the weights of the normal distances, thus favoring the nearby points close to the tangent plane where the points lie on the same side of the edge around query point $p$. However, due to the uneven density of point surfaces caused by the process of ICP, a voxel downsampling is necessary to execute with a fixed voxel size given by multiple of the average distance of the point cloud.

%The bilateral filter algorithm can better reduce the outliers among point cloud data and can help preserve the important geometric features of the welding workpiece surface. Therefore, this technique is used in this phase of the study to implement the filtering operation of the point cloud.

\begin{figure}[b]
	\centering
	\includegraphics[width=\columnwidth]{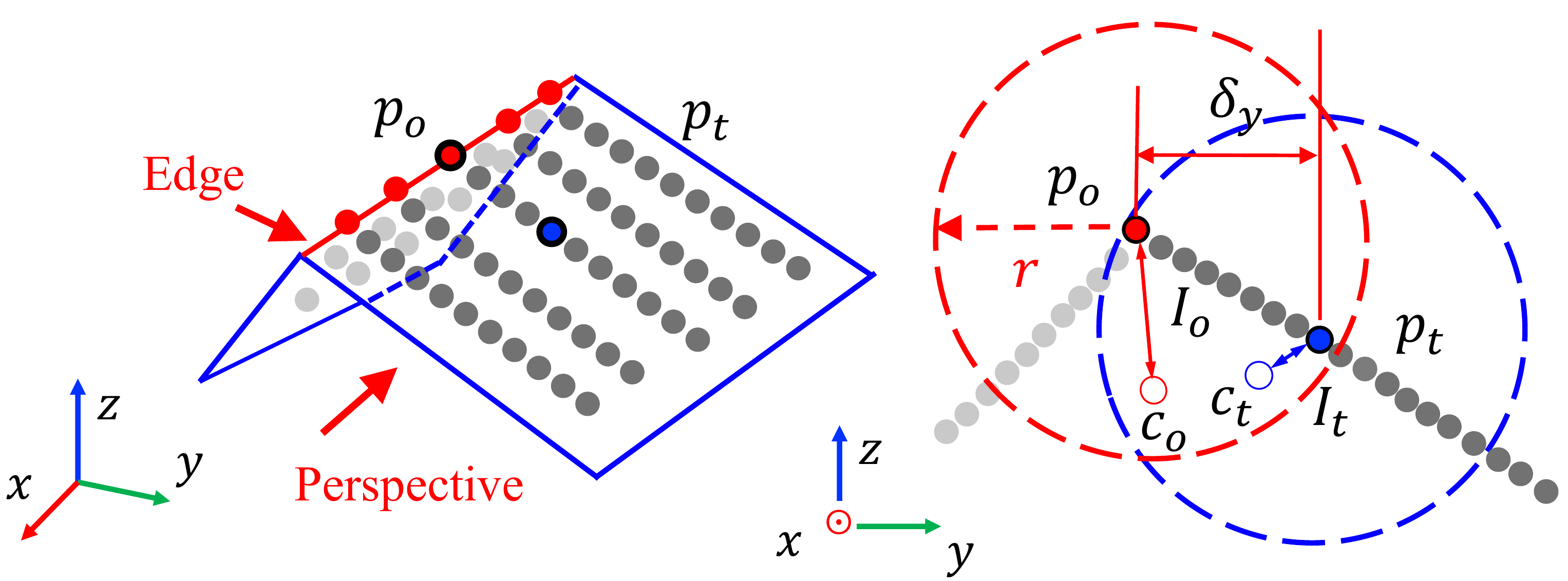}
	\caption{Conceptual representation of the edge intensity and gradient along the Y-axis for current point $p_o$, in which edge intensity $I_o$ with respect to $p_o$ is higher than $I_t$ with respect to its neighbor $p_t$. Gradient $I_{o_y}$ is defined as the difference between $I_o$ and $I_t$ divided by the coordinate difference along the Y-axis.}
	\label{fig:edge_detect}
\end{figure}
\subsection{Welding Seam Extraction}
In extracting the welding seam for various workpieces, a generic seam detection algorithm is proposed on basis of the gradients of the edge intensity. As the point density of a reconstructed workpiece model varies, few points of a neighborhood exist within a fixed range. The establishment of a minimum point $m$ will provide a rational neighborhood to reduce the effect on points density. Therefore, for a given point $p \in \mathcal{P}$, its neighboring point $q$ can be defined as:
\begin{equation}
 \mathcal{N}_{r}(p)= \left\{q \in \mathcal{P} \mid \|q-p\|_{2}<r \cup|\mathcal{N}| \geq m \right\}	
\end{equation}
where $r$ is the fixed radius, which can adaptively adjust to guarantee that no less than $m$ points will exist in the neighborhood of $p$. As shown in Fig. \ref{fig:edge_detect}, the normalized displacement between original point $p_o$ and its local geometric centroid $c_o$ can be defined as the edge intensity of $p_o$, which is calculated by:
\begin{equation}
	I_{o}=\frac{1}{r}\left\|p_o-\frac{1}{n} \sum_{i=0}^{i=n} q_{i}\right\|
\end{equation}
where $\frac{1}{n} \sum_{i=0}^{i=n} q_{i}$ denotes the geometric centroid for all the neighboring points corresponding to original point $p_o$. Within the edge-free areas, the intensity remains low, but the value will increase when approaching edges and reaches a local maximum in the intensity space at edges. Fig. \ref{fig:edge_detect} shows a side view of two intersecting point cloud surfaces as an example, in which the edge intensity $I_{o}$ of edge point $p_o$ is much higher compared with the intensity $I_{t}$ of its neighbor $p_t$ in non-edge region, as the distance between the local geometric centroid $c_o$ and $p_o$ is larger than the distance for $p_t$.

For 2D images, the gradients typically represent the direction in which the features of a pixel mostly change. The way of defining gradients for 3D point clouds can represent the variations between edge intensities within its neighborhood.
However, unlike the gradient decomposition in consecutive 2D pixels, gradients for a 3D point cloud can only be approximated due to the discrete points. Thus, we define the 3D gradient decompositions for a point $p_o$ as:
\begin{equation}
	\begin{aligned} 
	I_{k} & \approx \max _{i \in \mathcal{N}}\left[\left(I_{i}-I_{o}\right) \cdot 
	\frac{d_{i,o}}{\delta k}\right], k={x,y,z} \\ 
%	I_{y} & \approx \max _{i \in \mathcal{N}}\left[\left(I_{i}-I_{\circ}\right) \cdot 
%	\frac{d_{i,\circ}}{\delta y}\right] \\
%	I_{z} & \approx \max _{i \in \mathcal{N}}\left[\left(I_{i}-I_{\circ}\right) \cdot 
%	\frac{d_{i,\circ}}{\delta z}\right]
	\end{aligned}
\end{equation}
where $\delta x$, $\delta y$, and $\delta z$ are the coordinate differences in three axes, and 
$d_{i,o}$ represents the distance between a current point $p_o$ and its $i$-th neighboring point $q_{i}$. Fig. \ref{fig:edge_detect} shows an example of the point's gradient along Y axis.

A 3D offset function $F(x, y, z)$ for point clouds analoyous with the decomposition model for a small offset between pixels is presented and approximated by a Taylor series expansion. For a given point $p_o$, the offset function $F(x_o, y_o, z_o)$ can be expressed as a multiplication of a small offset $(\delta x, \delta y, \delta z)$ and a symmetric matrix $\mathbf{H}$, in which $I_{x}$, $I_{y}$, and $I_{z}$ can be calculated by:
\begin{equation}
	\begin{aligned} F(x_o, y_o, z_o) &= \sum_{\delta x, \delta y, \delta z} \left[ I_{x_o+\delta x, y_o+\delta y, z_o+\delta z} - I_{x, y, z} \right]^{2} \\ &=\sum_{\delta x, \delta y, \delta z} \left[ \begin{array}{c}  \delta x \cdot I_{x}+\delta y \cdot I_{y}+ \delta z \cdot I_{z} \\ +  O\left(\delta x^{2}, \delta y^{2}, \delta z^{2}\right)\end{array} \right]^{2}  \\ & \approx(\delta x, \delta y, \delta z) \mathbf{H}(\delta x, \delta y, \delta z)^{\top} \end{aligned}
\end{equation}
\vspace{-0.2cm}
\begin{equation}
	\mathbf{H}=\left[\begin{array}{ccc}I_{x}^{2} & I_{x} I_{y} & I_{x} I_{z} \\ I_{y} I_{x} & I_{y}^{2} & I_{y} I_{z} \\ I_{z} I_{x} & I_{z} I_{y} & I_{z}^{2}\end{array}\right]
\label{equ:14}
\end{equation}
where the distribution of the gradients for the local neighborhood is depicted by the eigenvalues of $\mathbf{H}$. 
Therefore, the intensity of gradient changes can be detected by comparing the difference of eigenvalues. For example, a high eigenvalue reveals a large shift in gradient along the related eigenvector, and this rule is widely used to classify the extracted features for 2D pixels. 
In extracting edges in point clouds, two observations will be detected, namely, 1) the co-existence of two high eigenvalues, which can indicate an edge lying on an intersection between two point cloud surfaces, and 2) a single existence of a high eigenvalue, which can reveal a boundary of the point cloud.
In the context of welding applications, both of the situations will be considered because the welding seam only lies on the intersection between two surfaces or the boundary between different workpieces. 
Additionally, as "high" eigenvalues with a threshold are hard to describe, the ratios between them are measured to identify a threshold $t$ instead of solely using the magnitude of eigenvalues.
Most importantly, this ratio calculation scheme does not rely on a time-consuming SVD.

To avoid the calculation of eigenvalues, we use $\lambda_1 = \alpha \lambda_3$ and $\lambda_2 = \beta \lambda_3$ $(\alpha \geq \beta \geq 1)$ to represent the three sorted eigenvalues $(\lambda_{1} \geq \lambda_2 \geq \lambda_3)$. Subsequently, the ratio between trace $\operatorname{Tr}(\mathbf{H})$ and determinant $\operatorname{Det}(\mathbf{H})$ can be calculated using Eq. \ref{equ:15}. As the ratio increases with larger $\alpha$ and $\beta$, we can set a threshold $t$ to guarantee the existence of large eigenvalues:
\begin{equation}
	\frac{\operatorname{Tr}(\mathbf{H})^{3}}{\operatorname{Det}(\mathbf{H})}=\frac{\left(\lambda_{1}+\lambda_{2}+\lambda_{3}\right)^{3}}{\lambda_{1} \lambda_{2} \lambda_{3}}=\frac{\left(\alpha+\beta+1\right)^{3}}{\alpha \beta} \geq t
	\label{equ:15}
\end{equation}

\begin{equation}
	G(o, i)=\frac{1}{\delta \sqrt{2 \pi}} e^{-\frac{\left\|p_{o}-q_{i}\right\|^{2}}{2 \delta^{2}}}, \quad q_{i} \in \mathcal{N}
\label{equ:16}
\end{equation}
Then, Eq. (\ref{equ:16}) is used to smoothen $\mathbf{H}$ and ensure $\operatorname{Det}(\mathbf{H})$ is a non-zero, in which $\sigma$ is the standard deviation of the distance between current $p_o$ and its neighborhood $\mathcal{N}_{r}(p_o)$. Finally, a set of  edge points $\{p^{(e)}_1, p^{(e)}_2, ..., p^{(e)}_n\}$ is generated.

\subsection{Welding Path Optimization}
On the basis of the extracted edge points, $p^{(w)}_i = p^{(e)}_i + d*\mathbf{n}_{p^{(e)}_i}$ is used to refine the edge point to welding path point for guiding the welding torch, where $d$ is the optimal distance to weld a workpiece suggested by different sets of welding equipment. 
\begin{figure}[t]
	\vspace{0.2cm}
	\centering
	\includegraphics[width=\columnwidth]{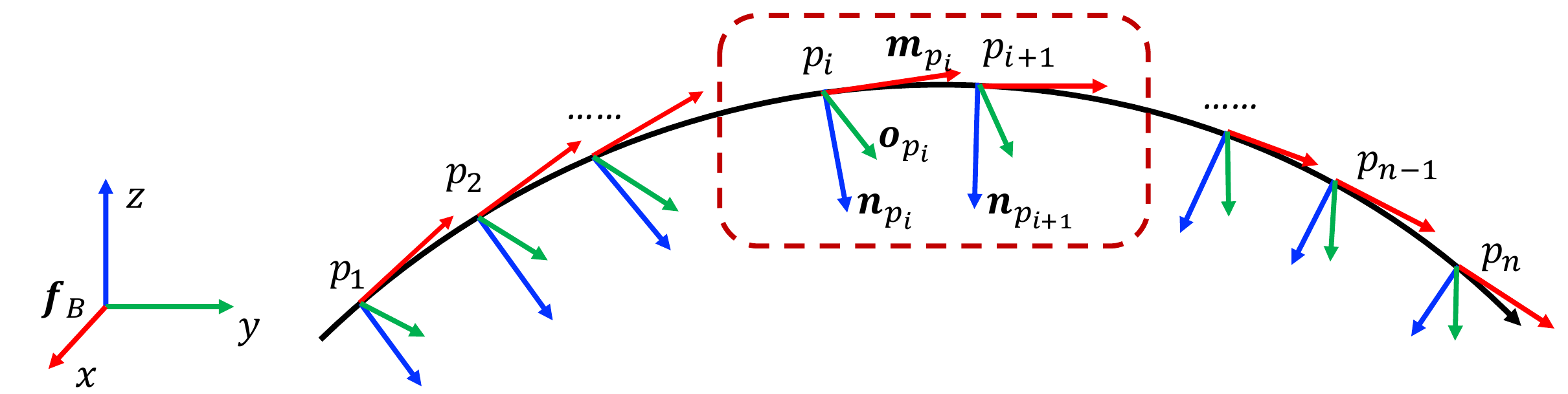}
	\caption{Determination of three unit orthogonal basis vectors of the 3-DOF rotation for each welding seam pose.
	}
	\label{fig:poseEstimate}
%	\vspace{-0.2cm}
\end{figure}
Once the 3-DOF translation parameters $\{p^{(w)}_1, p^{(w)}_2, ..., p^{(w)}_n\}$ for welding path points are determined, 3-DOF rotation parameters are necessary to form a complete welding path. The computation of rotation parameters needs to define three unit orthogonal basis vectors with respect to the robot base frame. As shown in Fig. \ref{fig:poseEstimate}, the first vector is defined as the pointing vector of the welding torch, which equals to the normal vector of the surrounding welding region point cloud. 
This approach aligns with the basic requirement of keeping the welding torch perpendicular to the welding region. The second vector is the moving direction of the welding torch, which is defined as the vector $\mathbf{v}_m$ between two nearest path points. Furthermore, to reduce the disturbance of the moving direction, its final moving vector is defined as the average vector of the $m$ nearest moving vectors, i.e., $\mathbf m_i  = (\sum^{i+\frac{m}{2}}_{i-\frac{m}{2}} \mathbf{v}_{m_i} ) / {m}$. According to the right-hand rule, the third vector is computed via cross-multiplication, i.e.,  $\mathbf o_i  = \mathbf  m_i  \times \mathbf n_i $. Therefore, each path point has its own coordinate that is formed by $(\mathbf o_i ,\mathbf m_i, \mathbf n_i)^T$.

In view of achieving superior welding motion performance, increasing the density of path points is necessary to support the robotic manipulator in smoothly executing the path-tracking motion \cite{rick2020}. A linear path interpolation method is designed from two aspects, namely, 3D translation interpolation and 3D rotation interpolation. In particular, a number of $\rho$ points is interpolated between $p^{(w)}_i$ and $p^{(w)}_{i+1}$ and the $j$-th interpolation point is defined as:
\begin{equation}\label{equa:jthpoint}
	\begin{aligned}
		p^{(w_i)}_j &= p^{(w)}_i + j \cdot \frac{p^{(w)}_{i+1} - p^{(w)}_{i}}{\rho}
	\end{aligned}  
\end{equation}
Then, each initial path point $p^{(w)}_{i}$ owns its rotation vector $\mathbf{v}^{(w)}_i$ and the $j$-th rotation vector is interpolated between $\mathbf{v}^{(w)}_i$ and $\mathbf{v}^{(w)}_{i+1}$, which is computed with the spherical--linear rotation interpolation method:
\begin{equation}\label{equa:rotationinter}
	\begin{aligned}
		\mathbf{v}^{(w_i)}_j = 
%		\text{slerp}(\mathbf{v}_i, \mathbf{v}_{i+1}, \alpha) 
\frac{ \sin((1-\alpha)\theta) \mathbf{v}^{(w)}_i + \sin{(\alpha\theta)} \mathbf{v}^{(w)}_{i+1}} {\sin \theta}
	\end{aligned}
\end{equation}
where $\alpha = j / \rho$ and $\theta = \arccos(\mathbf{v}^{(w)}_i, \mathbf{v}^{(w)}_{i+1})$. Finally, a dense and even 6-DOF path for welding is planned.

\section{Results}
\begin{figure}[htbp]
	\centering
	\includegraphics[width=\columnwidth]{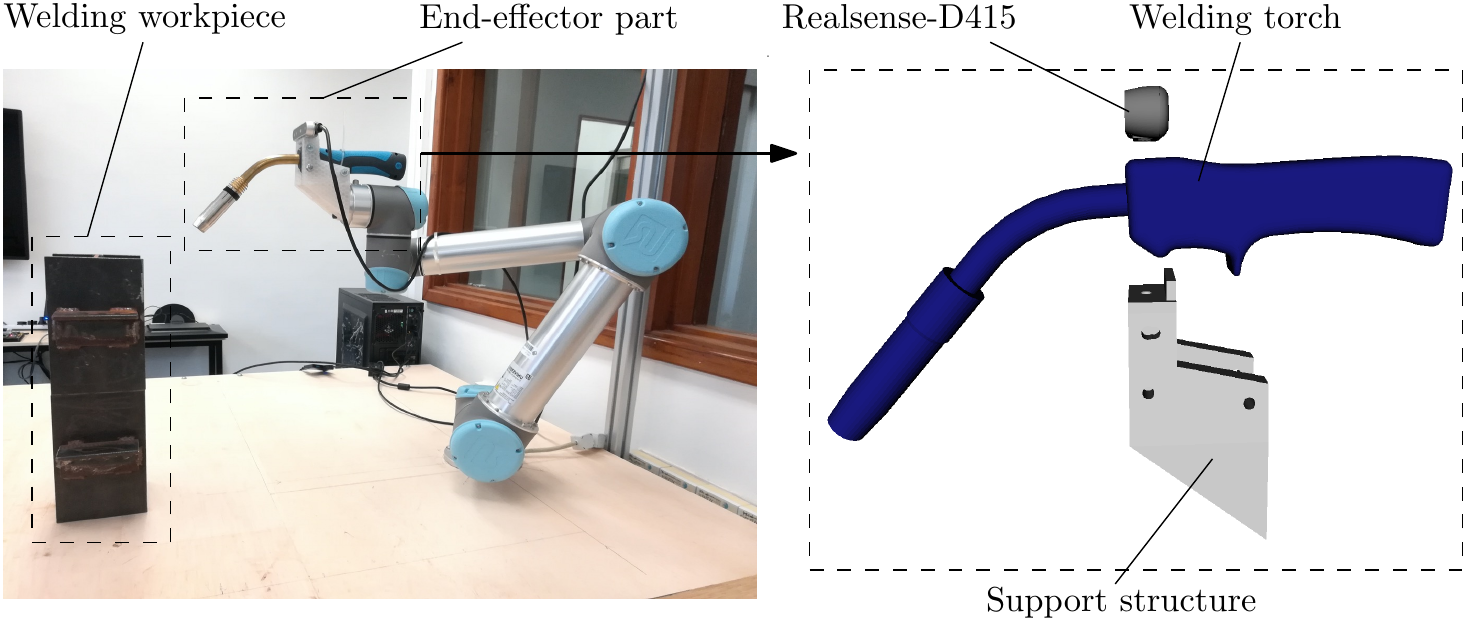}
	\caption{(Left) Experimental welding platform. (Right) End-effector of the proposed welding robotic system comprising an RGB-D camera and welding torch. The camera is positioned above the welding torch.}
	\label{fig10:experiment-setup}
\end{figure}

In evaluating the performance of the proposed welding path planning method, we built a robotic system based on UR5 and tested it on several workpieces. To verify the robustness, two complicated types (cylinder and Y-shaped workpieces) as shown in Fig. \ref{fig12:resutls} are selected in each following step of the entire planning process for illustration.
Fig. \ref{fig10:experiment-setup} has previously shown the experimental platform and the RGB-D sensor adopted in this work. The result of conventional teach--playback robotic path planning method, which is regarded the benchmark of welding path planning, is also obtained to compare with our proposed approach.

\subsection{Gesture-Guide Scanning}
\begin{figure}[htb]
	\centering
	\includegraphics[width=\columnwidth]{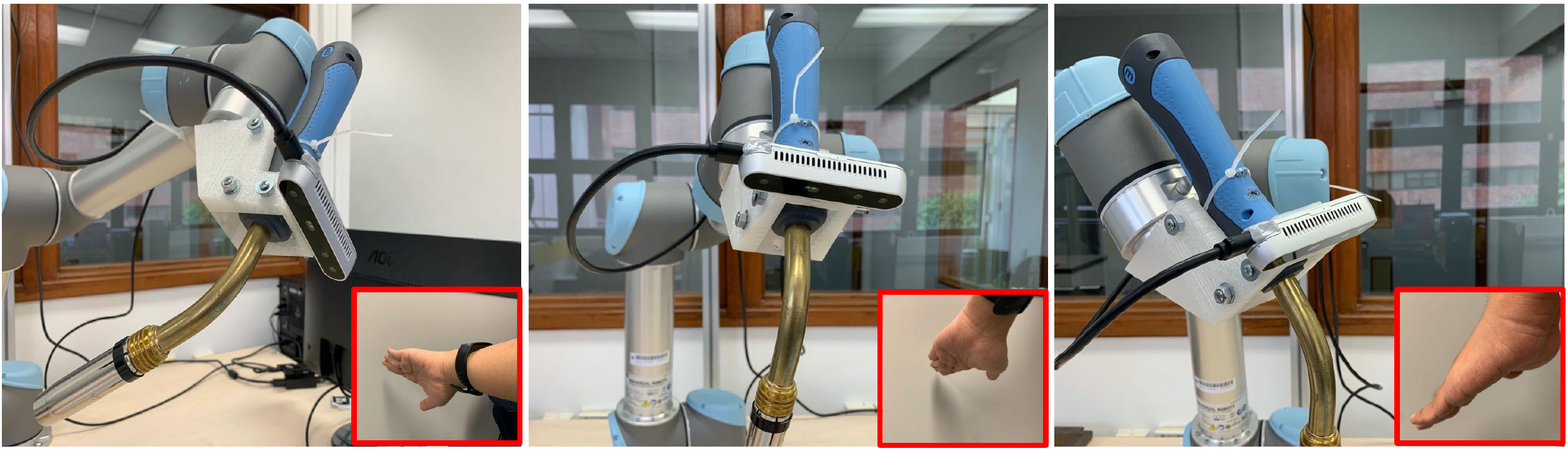}
	\caption{Results of the robotic scanning process for the workpiece with differently performed hand gestures.}
	\label{fig:interaction}
\end{figure}

In intuitively scanning the workpiece with bare hands, a gesture interaction is implemented via leap motion Python SDK ver. 3.2 and Python library for UR robots, namely, URX. With the leap motion SDK, the hand's palm position $^{LM}p_h$ with respect to the gesture sensor and angles of the raw, pitch and yaw $(\alpha, \beta, \gamma)$ for the left hand can be extracted in each captured frame. The interaction box is designed as a spatial cube with a length of $10 cm$ edge, which is appropriate for the interactions in this study. The pose in UR5 is represented by $(x, y, z, R_x, R_y, R_z)$, while an API ($scipy.spatial.transform$) is used to convert R-P-Y angles into a rotation vector. During interactions, the robotic manipulator only executes six pure linear motions and one rotation which are both based on hand gestures. Precision can be ensured theoretically with a given sufficient time periods. As shown in Fig. \ref{fig:interaction}, the approach direction of the RGB-D camera is fully aligned with the hand's normal. 
Consequently, the interaction time is the only key metric needed to measure this process. The interaction time is defined as the time period with respect to the time when the sensor receives the gesture to the camera, the robot starts to perform corresponding motions. The average interaction time of each motion is measured through 100 gesture actions. The results show that the rotation motion needs more execution time, compared with linear motions. Linear interaction time is all approximately $0.05 s$, which is adequate for a normal interaction. By contrast, the average interaction time for rotations is approximately $0.9 s$, depending on the complexity of the transformation.

\begin{figure}[b]
	\centering
	\includegraphics[width=\columnwidth, height=0.4\columnwidth]{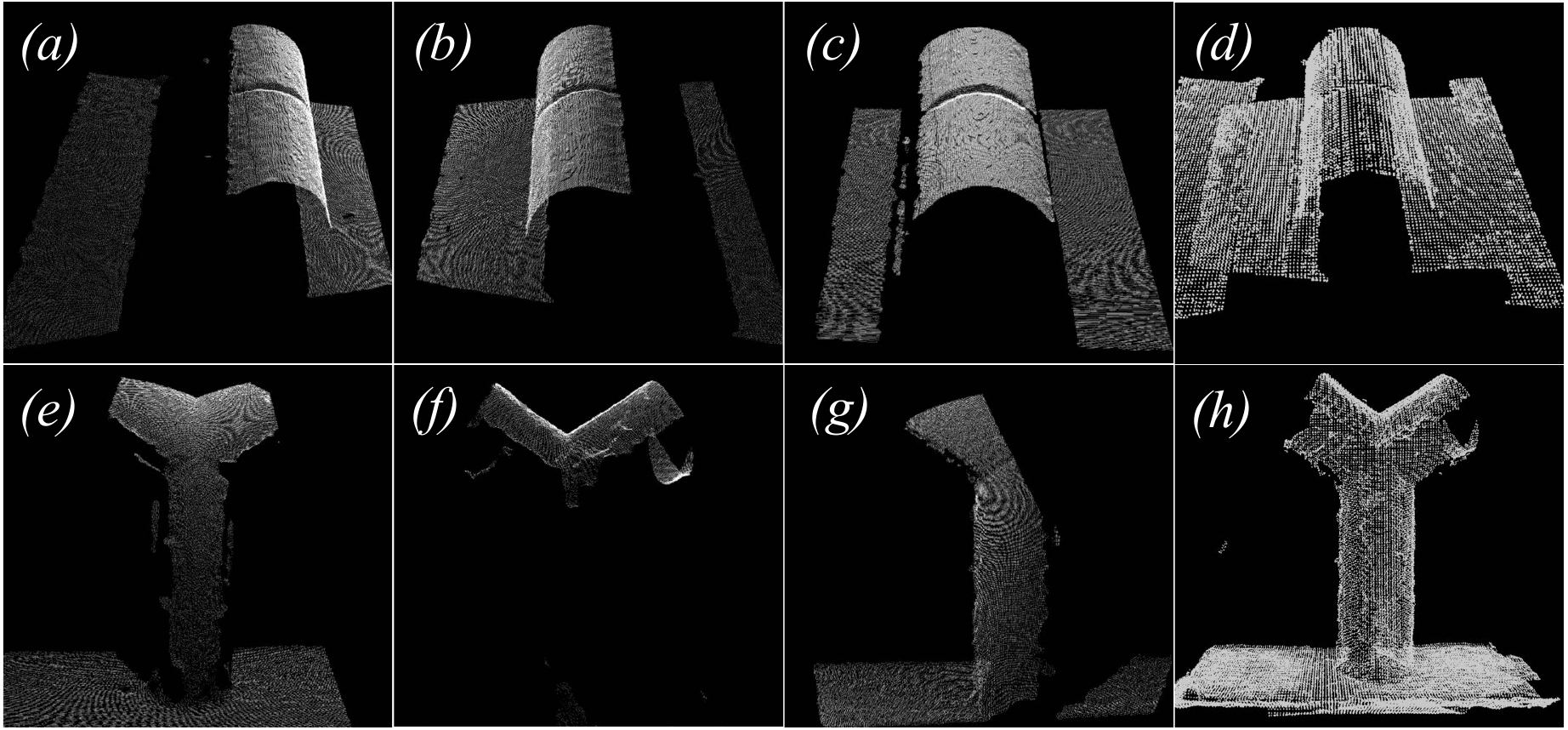}
	\caption{Model reconstruction. 
	(a)--(c), (e)--(g) Partial point clouds from different views.
	(d, h) Merge partial point clouds to form a completed model.
	}	
	\label{fig:reconstruction}
\end{figure}
\begin{figure*}[htb]
\vspace{0.2cm}
	\centering
	\includegraphics[width=\columnwidth * 2
	]{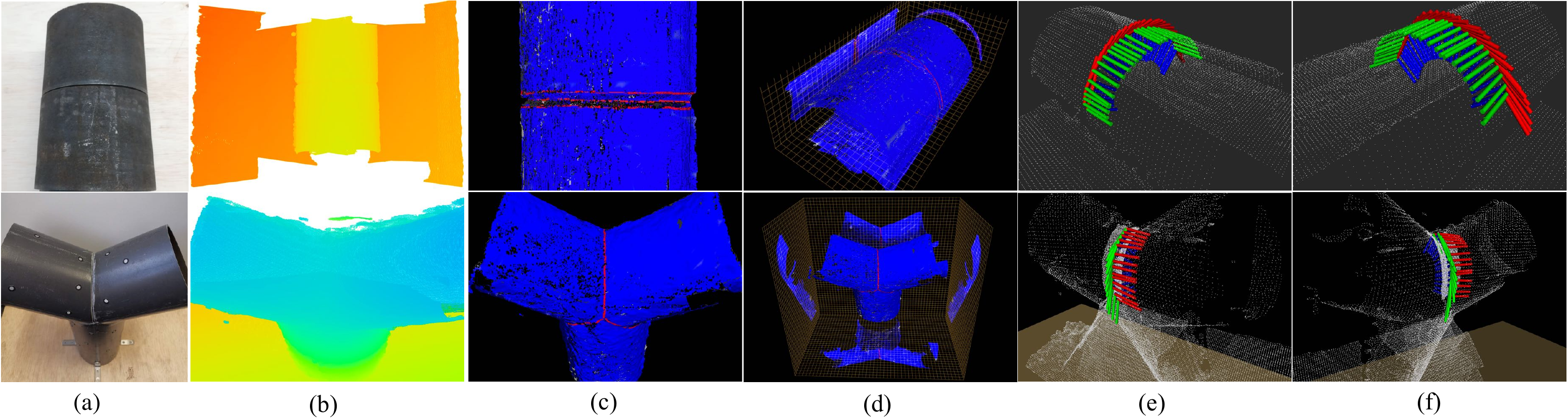}
	\caption{Results of different processes in the proposed system.
	(a) Two actual welding metal workpieces, cylinder and Y-shape.
	(b) Reconstructed models of workpieces with a depth map.
	(c) Welding seam extracted with the edge intensity gradient-based algorithm.
	(d) Rasterized views of extracted edge points.
	(e, f) 6-DOF welding path after interpolation.
	}
	\label{fig12:resutls}
	\vspace{-0.2cm}
\end{figure*}
\subsection{Workpiece Reconstruction}
\begin{figure}[htb]
	\centering
	\includegraphics[width=0.69\columnwidth]{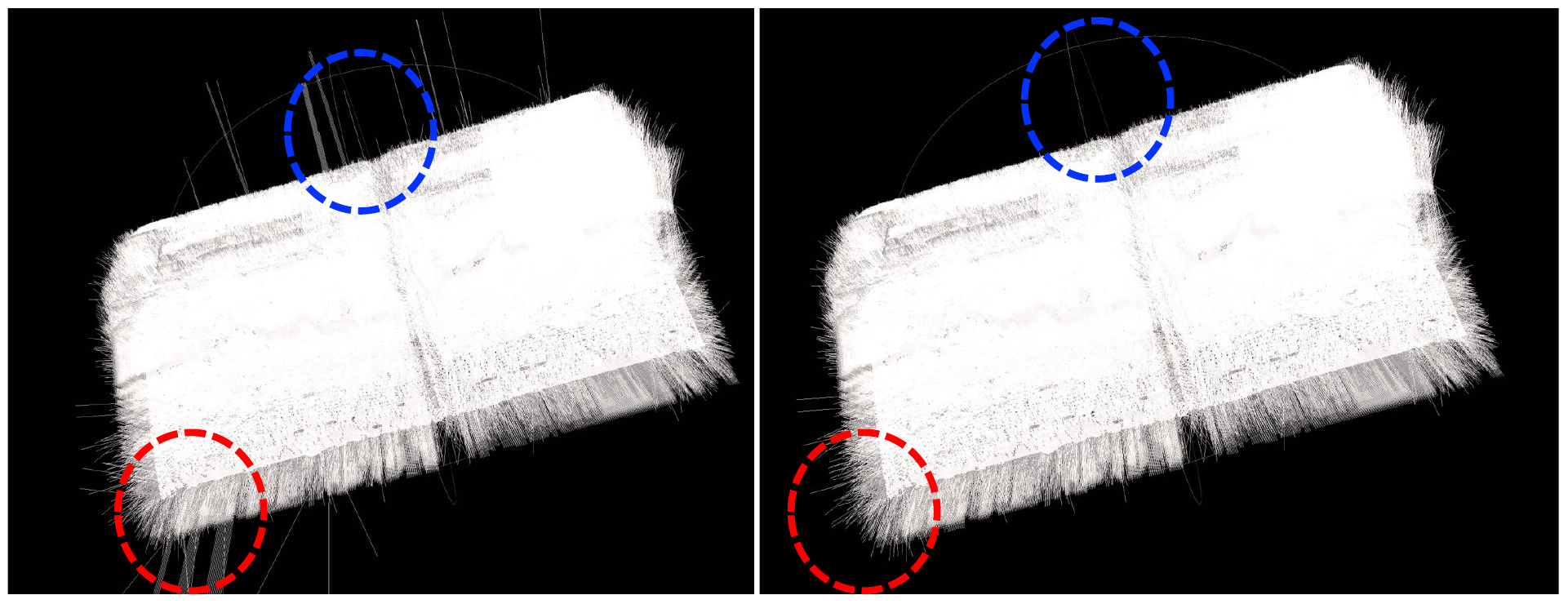}
	\includegraphics[width=0.29\columnwidth]{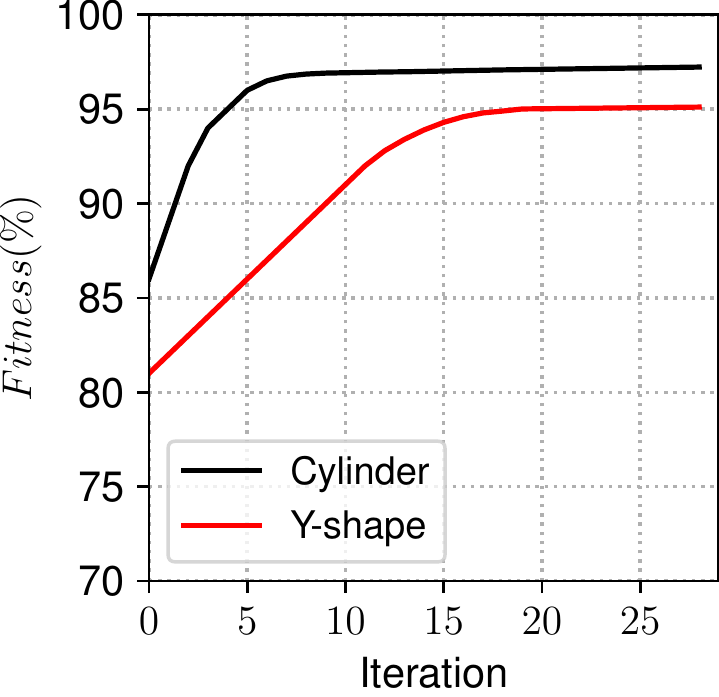}
	\caption{(Left) Comparative results of pre and post-denoising with the bilateral filter.
	(Right) Fitness score of two workpieces with different iterations.}
	\label{fig:filterRes}
\end{figure}

As shown in Fig. \ref{fig:reconstruction}, partial point clouds from different perspectives are generated during the scanning process. 
The camera pose with respect to the world frame is calculated using a forward kinematics module provided by URX and a fixed transformation from the camera to the robot end-effector. On the basis of this information, a preliminary 3D model of the workpieces are reconstructed, as shown in Figs. \ref{fig:reconstruction}(d) and (h). In measuring the reconstruction performance, a metric of the fitness score is computed as follows:

%squared error of Euclidean distance
%\mathbf{M}}_{\mathrm{opt}

\begin{equation}
%	P_{i} \in \mathcal{N} \Leftrightarrow\left\|P_{c}-P_{i}\right\| \leq R \cup|\mathcal{N}| \geq M
	S(\mathbf{M})= \frac{|\mathcal{P}_{m}|}
	{|\mathcal{P}_{t}|}, \left\{ m_{i} \in \mathcal{P}_{m} \mid
	\left\|t_{i}-\mathbf{M} m_{i}\right\|^{2} <d \right\}
\end{equation}
where $t_{i}$ and $m_{i}$ are the target point and the model point, respectively, and $\mathbf{M}$ is the iteratively optimized transformation. Therefore, this score can measure the ratio of successfully aligned points. As depicted by the trend in Fig. \ref{fig:filterRes}, the fitness score of the cylinder piece represented by the red spline is higher than the other Y-shaped piece labeled with the black spline throughout the entire ICP process. 
Before the ICP, the model of the cylinder piece was 86\%, and that of the Y-shaped piece was 81\%, but these values are not acceptable in high-precision welding. During the ICP process, the cylinder piece quickly reached the convergence when the iteration equalled 5, while the Y-shaped piece needed 15 iterations to reach the convergence. This trend can be explained by the Y-shaped piece that has a more complex geometrical surface than the cylinder piece.

\subsection{Welding Path Detection}
\begin{figure}[htbp]
	\centering
	\includegraphics[width=\columnwidth]{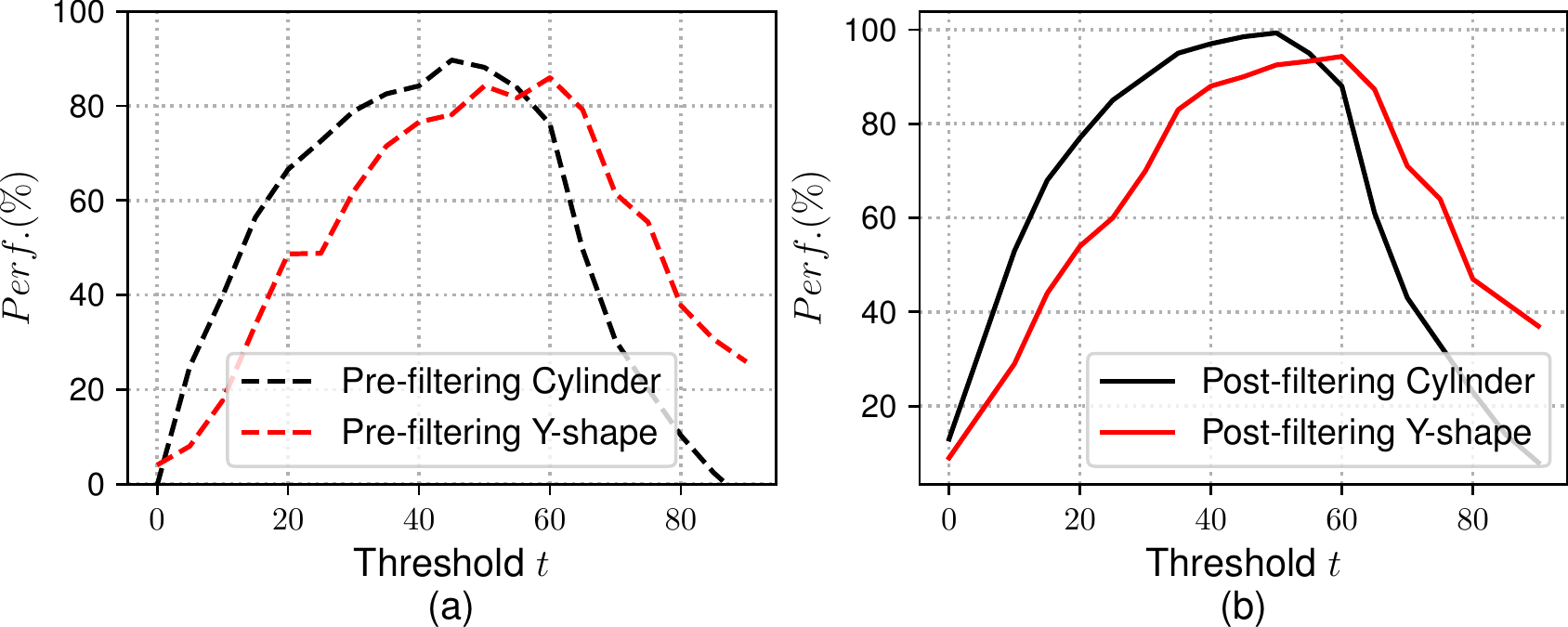}
	\caption{Pre- and post-performance scores under different thresholds.}	
	\label{fig:threshold}
\end{figure}

\begin{figure}[htb]
	\centering
	\includegraphics[width=\columnwidth]{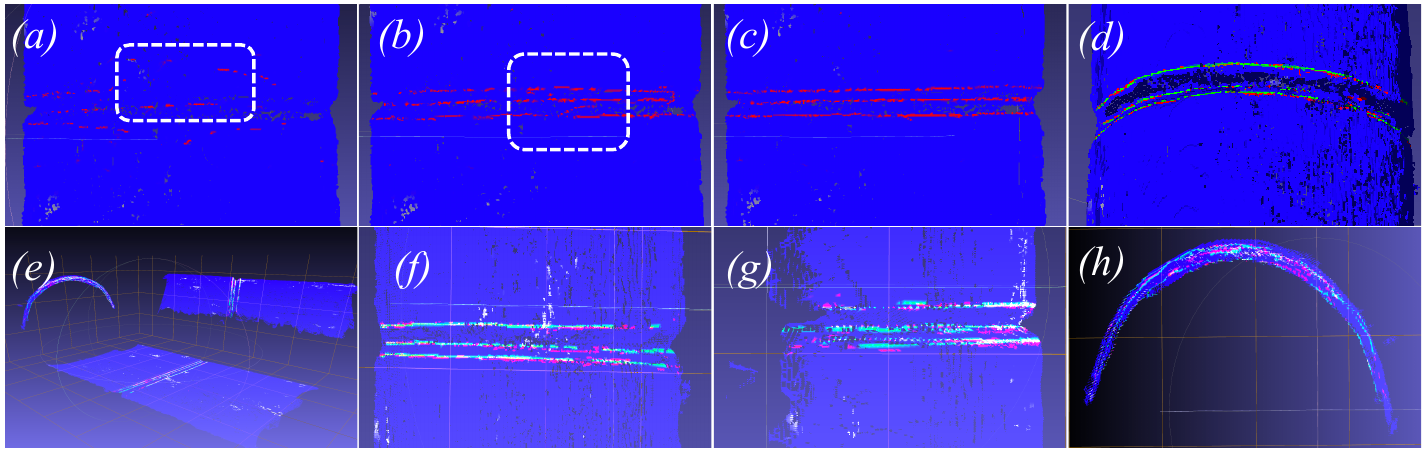}
	\caption{(a)--(c) Overall performance with $t = 10, 70, 50$
	(d) Extracted edge points and ground truth edge points.
	(e)--(h) Rasterization to measure the precision of the generated welding path.
	}	
	\label{fig:Raster}
\end{figure}

After rasterization, the harmonic mean of accuracy is taken to measure the overall performance. The mean is defined as:
\begin{equation}
F_1 = \frac{3}{a_{xy}^{-1}+a_{xz}^{-1}+a_{yz}^{-1}}
\end{equation}
where $a_{xy}^{-1}$, $a_{xz}^{-1}$ and $a_{yz}^{-1}$ denote the accuracy calculated for the $xy$, $xz$, and $yz$ planes after rasterization. The corresponding rasterized point clouds are shown in Figs. \ref{fig:resutls} (e)-(h). 
Each detected edge point is displayed in red, and the ground truth path point is manually marked in green based on point cloud generated by a high-precision 3D scanner. 
The overall performance calculated on the basis of the coverage of the red edge point over the ground truth path rasterized on different planes.
The effect of the bilateral filter on the welding path detection also needs to be determined. Figs. \ref{fig:threshold} (a) and (b) show the overall pre- and post-filtering performances of the two selected workpieces, respectively.
The figures show that the performances of the two workpieces both have increased after filtering. This trend is due to the filtering, which reduces the noise and smoothens the surface for the point cloud data. Fig. \ref{fig:filter} validates this difference based on the normal map of the pre- and post-filtering processes of the workpieces, especially on the corresponding regions surrounded by dashed circles. 
Besides, different predefined thresholds $t$ have an impact on the overall performance. Figs. \ref{fig:Raster}(a)-(c) show the extracted edges when $t$ is equal to $10$, $70$, and $50$, respectively. If the $t$ is too small or too large, then it will negatively affect the overall performance score. The cylinder has reached the peak when its threshold is near $53$, whereas the Y-shaped piece needs a slightly larger threshold of approximately  $60$ to attain the peak performance.

\subsection{Pose Optimization}
After optimization, the welding path planned by the teach--playback method is selected as the standard path, which is regarded as our ground truth for comparing the welding path generated by our framework. 
For each workpiece, two individual paths are generated, and the root mean square error (RMSE) of the two paths is used to evaluate both the translation and rotation accuracy of the proposed method.
The RMSE is defined as:
\begin{equation}\label{RMSE}
	\begin{aligned}
		RMSE(m, g) = \sqrt{ \frac{1}{n} \sum^n_{i=1} (m_i - g_i)^2}
	\end{aligned}
\end{equation}
where $\{m_i\}$ represents the generated welding path by our model, $\{g_i\}$ is the ground truth of the welding path, and $n$ denotes the number of path points. The RMSE results of the two workpieces are shown in Table \ref{tab:rmse}, which reveals acceptable accuracy for both translation and rotation.
\begin{table}[h]
\vspace{0.25cm}
    \begin{center}
    \caption{Results of Pose Accuracy for each workpiece} \label{tab:rmse}
        \begin{tabular}{lcccc}
            \toprule
            Types &$n_d$ &$n_o$ &$RMSE_{p}$ & $RMSE_{\theta}$\\             
            \midrule       
            Cylinder &78 & 300  & $0.7mm$  & $1.5^{\circ}$ \\            
            Y-shaped  &65 & 300  & $0.8mm$  & $1.3^{\circ}$ \\
            \bottomrule
        \end{tabular}
    \vspace{5pt}
    \newline
	$n_d$ = The number of detected path points \\
	$n_o$ = The number of points after optimization.
    \end{center}
\end{table}
\subsection{Comparison with Existing Methods}
Compared with previous studies on groove detection, thresholding eigenvalue of gradient makes system performance less vulnerable to noise in complex environments and thus our system can be flexibly applied to a wide range of workpiece. This is supported by sufficient experiments from plane placed at various poses to complex 3d shapes. 
However, the output welding path is computed based on previous downsampled point cloud which may fail to maintain concrete local geometric information, especially for points at the corners.
Accuracy is another weakness of the system as the standard value given by other studies \cite{xu2012real, fan2019initial, diao2017passive, nele2013image} is under 0.5 mm.
% The gestured based scanning leaves more space for human-robot collaboration and can possibly be applied in other settings to substitute manual control or pre-programmed positions. 
% In this way, the initial detection efficiency can improve substantially. Combined with a broad view vision sensor, the system can reconstruct free form workpieces and plan the path at a competitively fast speed. 
Instead of using manual controls or pre-programming poses for positioning the camera, we adopt a gestured based interaction for scanning. Combined with a broad view vision sensor, the system can reconstruct free form workpieces and plan the path at a competitively fast speed. 
We measured the runtime for the whole system on a 500 mm straight V-type butt joint as this type is commonly presented in the literature. The full workpiece can be reconstructed from one capture. The average time for 20 runs is taken for comparison and the detection speed defined by length of seam over runtime is calculated. The results listed in Table \ref{tab:runtime} show that our system outputs consistent results across different settings and is significantly efficient for seam detection.

\begin{table}[htbp]
    \begin{center}
    \caption{Runtime on V-type butt joint welding path planning} \label{tab:runtime}
        \begin{tabular}{lcccc}
            \toprule
            System &Joint Pose &Method &Runtime(ms) &Speed(mm/s)\\          
            \midrule       
            Ours &flat &Point cloud &3691 &135 \\ 
            Ours &tilted &Point cloud &3891 &129 \\ 
            Ours &horizontal &Point cloud &3600 &139 \\ 
            Ours &vertical &Point cloud &4100 &122 \\ 
            Ref\cite{manorathna2014feature}  &flat &Laser vision &\textbackslash &5 \\  
            Ref\cite{yang2020novel} &flat &Point cloud &4913 &\textbackslash \\ 
            \bottomrule
        \end{tabular}
    \end{center}
\end{table}

\section{Conclusion}
This article presents a sensor-guided welding robotic system. 
Specifically, an interactive algorithm based on gesture is first proposed to scan the workpiece in an intuitive manner. 
Then, the system reconstructs a 3D point cloud model of the target by a linear ICP algorithm. The precision and speed of the reconstruction is ensured by taking the pose obtained from robotic kinematics as the initial setting in ICP.
Next, a bilateral filter is implemented to denoise point cloud while maintaining edge information and the welding seam joints is robustly identified using a gradient-based edge detection approach.
Based on the detected seam, a smooth 6-DOF path is computed for execution.
Experiments on various geometric shapes have validated the effectiveness of the proposed system.
In the future, a higher precision depth camera will be used to improve the accuracy of model reconstruction. Additionally, we may introduce other new interaction techniques to assist robotic welding tasks.

\bibliographystyle{IEEEtran}
\bibliography{root}

\begin{thebibliography}{10}
\providecommand{\url}[1]{#1}
\csname url@rmstyle\endcsname
\providecommand{\newblock}{\relax}
\providecommand{\bibinfo}[2]{#2}
\providecommand\BIBentrySTDinterwordspacing{\spaceskip=0pt\relax}
\providecommand\BIBentryALTinterwordstretchfactor{4}
\providecommand\BIBentryALTinterwordspacing{\spaceskip=\fontdimen2\font plus
\BIBentryALTinterwordstretchfactor\fontdimen3\font minus
  \fontdimen4\font\relax}
\providecommand\BIBforeignlanguage[2]{{%
\expandafter\ifx\csname l@#1\endcsname\relax
\typeout{** WARNING: IEEEtran.bst: No hyphenation pattern has been}%
\typeout{** loaded for the language `#1'. Using the pattern for}%
\typeout{** the default language instead.}%
\else
\language=\csname l@#1\endcsname
\fi
#2}}

\bibitem{xu2012real}
Y.~Xu, H.~Yu, J.~Zhong, T.~Lin, and S.~Chen, ``Real-time seam tracking control
  technology during welding robot gtaw process based on passive vision
  sensor,'' \emph{J Mate Process Tech.}, vol. 212, no.~8, pp. 1654--1662, 2012.

\bibitem{fan2019initial}
J.~Fan, S.~Deng, F.~Jing, C.~Zhou, L.~Yang, T.~Long, and M.~Tan, ``An initial
  point alignment and seam-tracking system for narrow weld,'' \emph{IEEE Trans.
  Industr. Inform.}, vol.~16, no.~2, pp. 877--886, 2019.

\bibitem{diao2017passive}
C.~Diao, J.~Ding, S.~Williams, Y.~Zhao, \emph{et~al.}, ``A passive imaging
  system for geometry measurement for the plasma arc welding process,''
  \emph{IEEE Trans. Ind. Electron.}, vol.~64, no.~9, pp. 7201--7209, 2017.

\bibitem{li2017automatic}
X.~Li, X.~Li, S.~S. Ge, M.~O. Khyam, and C.~Luo, ``Automatic welding seam
  tracking and identification,'' \emph{IEEE Trans. Ind. Electron.}, vol.~64,
  no.~9, pp. 7261--7271, 2017.

\bibitem{shah2016review}
H.~N.~M. Shah, M.~Sulaiman, A.~Z. Shukor, M.~Rashid, and M.~Jamaluddin,
  ``Review paper on vision based identification, detection and tracking of weld
  seams path in welding robot environment,'' \emph{Mod. App. Sci.}, vol.~10,
  no.~2, pp. 83--89, 2016.

\bibitem{rout2019advances}
A.~Rout, B.~Deepak, and B.~Biswal, ``Advances in weld seam tracking techniques
  for robotic welding: A review,'' \emph{Robot. Cim-Int. Manuf.}, vol.~56, pp.
  12--37, 2019.

\bibitem{nele2013image}
L.~Nele, E.~Sarno, and A.~Keshari, ``An image acquisition system for real-time
  seam tracking,'' \emph{Int. J. Adv. Manuf. Tech.}, vol.~69, no. 9-12, pp.
  2099--2110, 2013.

\bibitem{ding2016line}
Y.~Ding, W.~Huang, and R.~Kovacevic, ``An on-line shape-matching weld seam
  tracking system,'' \emph{Robot. Cim-Int. Manuf.}, vol.~42, pp. 103--112,
  2016.

\bibitem{manorathna2014feature}
R.~Manorathna, P.~Phairatt, P.~Ogun, T.~Widjanarko, M.~Chamberlain, L.~Justham,
  S.~Marimuthu, and M.~R. Jackson, ``Feature extraction and tracking of a weld
  joint for adaptive robotic welding,'' in \emph{Int. Conf. Cont. Auto. Robot
  \& Vis.}, 2014, pp. 1368--1372.

\bibitem{zhang2018point}
L.~Zhang, Y.~Xu, S.~Du, W.~Zhao, Z.~Hou, and S.~Chen, ``Point cloud based
  three-dimensional reconstruction and identification of initial welding
  position,'' in \emph{Trans. Intel. Weld. Manuf.}, 2018, pp. 61--77.

\bibitem{ahmed2018edge}
S.~M. Ahmed, Y.~Z. Tan, \emph{et~al.}, ``Edge and corner detection for
  unorganized 3d point clouds with application to robotic welding,'' in
  \emph{{IEEE/RSJ Int. Conf. on Robots and Intelligent Systems}}, 2018, pp.
  7350--7355.

\bibitem{patil2019extraction}
V.~Patil, I.~Patil, \emph{et~al.}, ``Extraction of weld seam in 3d point clouds
  for real time welding using 5 dof robotic arm,'' in \emph{Int. Conf. Cont.,
  Auto. and Robot}, 2019, pp. 727--733.

\bibitem{jing2016rgb}
L.~Jing, J.~Fengshui, and L.~En, ``Rgb-d sensor-based auto path generation
  method for arc welding robot,'' in \emph{Chinese Control and Decision Conf.},
  2016, pp. 4390--4395.

\bibitem{zhang20193d}
K.~Zhang, M.~Yan, T.~Huang, J.~Zheng, and Z.~Li, ``3d reconstruction of complex
  spatial weld seam for autonomous welding by laser structured light
  scanning,'' \emph{J. Manuf. Process}, vol.~39, pp. 200--207, 2019.

\bibitem{low2004linear}
K.-L. Low, ``Linear least-squares optimization for point-to-plane icp surface
  registration,'' \emph{Chapel Hill, Univ. North Carolina}, vol.~4, no.~10, pp.
  1--3, 2004.

\bibitem{rick2020}
R.~Peng, D.~Navarro-Alarcon, \emph{et~al.}, ``A point cloud based method for
  automatic groove detection and trajectory planning of robotic arc welding
  tasks,'' \emph{IEEE Int. Conf. on Ubiquitous Robots}, pp. 1--7, 2020.

\bibitem{yang2020novel}
L.~Yang, Y.~Liu, J.~Peng, and Z.~Liang, ``A novel system for off-line 3d seam
  extraction and path planning based on point cloud segmentation for arc
  welding robot,'' \emph{Robot. and Cim-Int. Manuf.}, vol.~64, p. 101929, 2020.

\end{thebibliography}
% For submission, pls remove .bib
\end{document}